\documentclass[letterpaper, 10 pt, conference]{ieeeconf}  

\IEEEoverridecommandlockouts

\usepackage{graphicx}
\usepackage{subfigure}
\usepackage{cite}
\usepackage{amsmath,amssymb,amsfonts}
\usepackage{tabularx,ragged2e,booktabs}
\usepackage{multirow}
\usepackage{tikz}
\usepackage{textcomp}
\usepackage{hyperref}
\usepackage{lipsum}
\usepackage{nicefrac}
\usepackage{pifont}
\newcommand{\cmark}{\ding{51}}%
\newcommand{\xmark}{\ding{55}}%
\usepackage{color}
\definecolor{mygreen}{RGB}{34, 139, 34}

\newcommand\copyrighttext{%
  \footnotesize This work has been submitted to the IEEE for possible publication. Copyright may be transferred without notice, after which this version may no longer be accessible.
  }
\newcommand\copyrightnotic{%
\begin{tikzpicture}[remember picture,overlay]
\node[anchor=south,yshift=10pt] at (current page.south) {\fbox{\parbox{\dimexpr\textwidth-\fboxsep-\fboxrule\relax}{\copyrighttext}}};
\end{tikzpicture}%
}

\title{\LARGE \bf
Multi-modal Streaming 3D Object Detection}

\author{Mazen Abdelfattah$^{1}$, Kaiwen Yuan$^{1}$, Z. Jane Wang$^{1}$, and Rabab Ward$^{1}$

\thanks{$^{1}$Mazen Abdelfattah, Kaiwen Yuan, Z. Jane Wang, and Rabab Ward are with the Department of Electrical and Computer Engineering, University of British Columbia, Vancouver, BC V6T1Z4, Canada (email: \{mazen,  kaiwen, zjanew, rababw\}@ece.ubc.ca;). \emph{Corresponding author: Mazen Abdelfattah.}}%
}

\begin{document}

\maketitle
\copyrightnotic
\thispagestyle{empty}
\pagestyle{empty}



\begin{abstract}

Modern autonomous vehicles rely heavily on mechanical LiDARs for perception. Current perception methods generally require $360^\circ$ point clouds, collected sequentially as the LiDAR scans the azimuth and acquires consecutive wedge-shaped slices. The acquisition latency of a full scan ($\sim 100ms$) may lead to outdated perception which is detrimental to safe operation. Recent streaming perception works proposed directly processing LiDAR slices and compensating for the narrow field of view (FOV) of a slice by reusing features from preceding slices. These works, however, are all based on a single modality and require past information which may be outdated. Meanwhile, images from high-frequency cameras can support streaming models as they provide a larger FoV compared to a LiDAR slice. However, this difference in FoV complicates sensor fusion. To address this research gap, we propose an innovative camera-LiDAR streaming 3D object detection framework that uses camera images instead of past LiDAR slices to provide an up-to-date, dense, and wide context for streaming perception. The proposed method outperforms prior streaming models on the challenging NuScenes benchmark. It also outperforms powerful full-scan detectors while being much faster. Our method is shown to be robust to missing camera images, narrow LiDAR slices, and small camera-LiDAR miscalibration. 

\end{abstract}


\vspace{-0.2cm}
\section{INTRODUCTION}
\label{sec:intro}
\begin{figure}[htbp]
\centering
\includegraphics[width=0.88\linewidth]{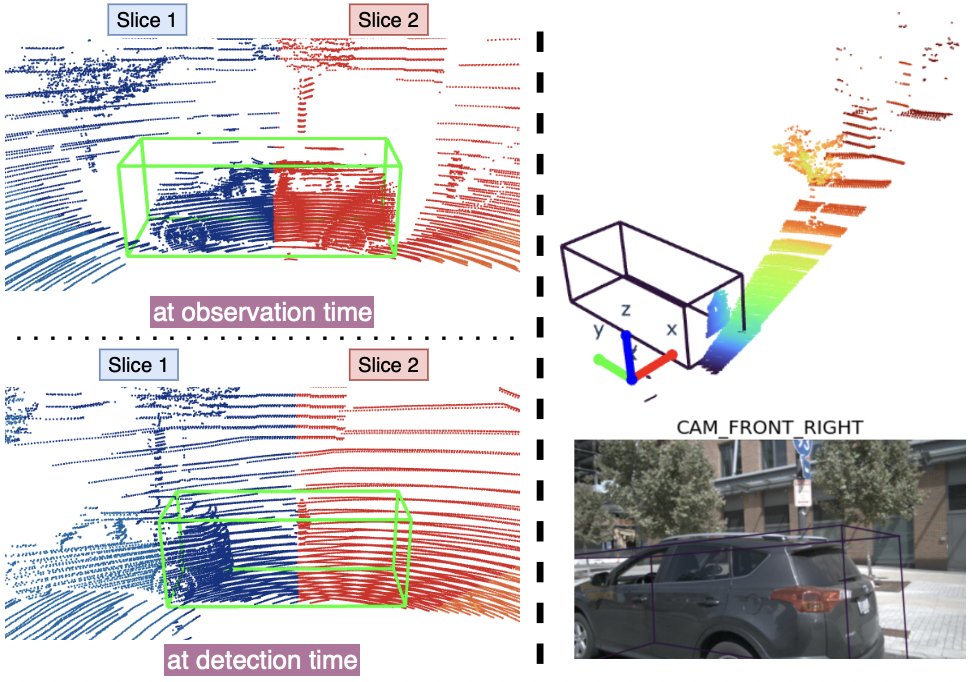}
\vspace{-0.2cm}
\caption{\hspace{-0.05cm}Left: The latency of waiting for a full LiDAR scan can lead to erroneous results for safety-critical tasks like detection. In a very dynamic scene, past slices may contain outdated information, and objects close to the ego-vehicle can be divided between slices, posing a challenge to streaming models. Right: Taking only \nicefrac{1}{32} of a full LiDAR scan, our model can accuratley detect heavily fragmented objects with the aid of the image input.}
\label{fig:delay}
\vspace{-0.6cm}
\end{figure}

Safe navigation for autonomous vehicles (AVs) requires minimal latency in perception, especially in dynamic environments. Dominant LiDAR-based 3D object detectors often rely on rotational LiDARs to provide a $360^\circ$ point cloud of the scene around an AV \cite{nuscenes,centerpoint, wang2021pointaugmenting}. A full LiDAR sweep is composed of a stream of slices/packets collected sequentially as the LiDAR scans the azimuth. These detectors must wait for a full LiDAR scan ($\sim 100 ms$) before inference which presents a bottleneck for perception latency. This delay can lead to outdated detections which is detrimental to safe autonomous driving \cite{stream, strobe}, as shown in Fig. \ref{fig:delay}.

Instead of waiting for a full scan, \cite{stream} proposed directly taking a slice (once it arrives) as input for detection models. While processing LiDAR slices reduces latency, the narrow field of view (FoV) of a slice results in less available context for the model to learn and infer from, thus resulting in lower accuracy. This is especially problematic for objects close to the ego-vehicle (the most safety-critical). A slice may only contain a fragment of an object, as shown in Fig. \ref{fig:delay}. To tackle this problem, all prior work on streaming 3D detection \cite{stream, strobe, polarstream} have utilized memory mechanisms and features from past slices to provide a global context.

The use of \emph{past features} is not ideal (especially in dynamic scenes) since it expands the current context with outdated information. A vehicle with a speed of 60 km/h moves nearly 1.7 m during a single scan by a 10 Hz LiDAR. An object of interest in previous slices may no longer be in the same position when processing the current slice due to the fast movement of objects, sudden occlusion, or a multitude of different factors inherent in a dynamic environment. These are issues that simple ego-motion compensation cannot fix, since it does not address the motion of other objects and it can be error prone \cite{pointmoseg}. Errors resulting from using past LiDAR features will aggregate and affect downstream tasks. 

This concern has not been addressed previously in streaming perception \cite{stream, strobe, polarstream}. One reason could be that the artificial experimental setup of prior studies relied on relatively static LiDAR scans (discussed in Sec. \ref{sec:rel}). Also, the use of features from past slices relies on two assumptions that may not hold in practice: 1) that slice inference time is shorter than the arrival rate of slices, and 2) that past slices truly represent the current world. If 1) fails, then a \emph{sequential} bottleneck arises where arriving slices wait for prior slices to be processed. If 2) fails, then we get outdated or misinformed perception.

\begin{figure*}[t]
 \center
 \vspace{-0.2cm}
  \includegraphics[width=0.9\linewidth]{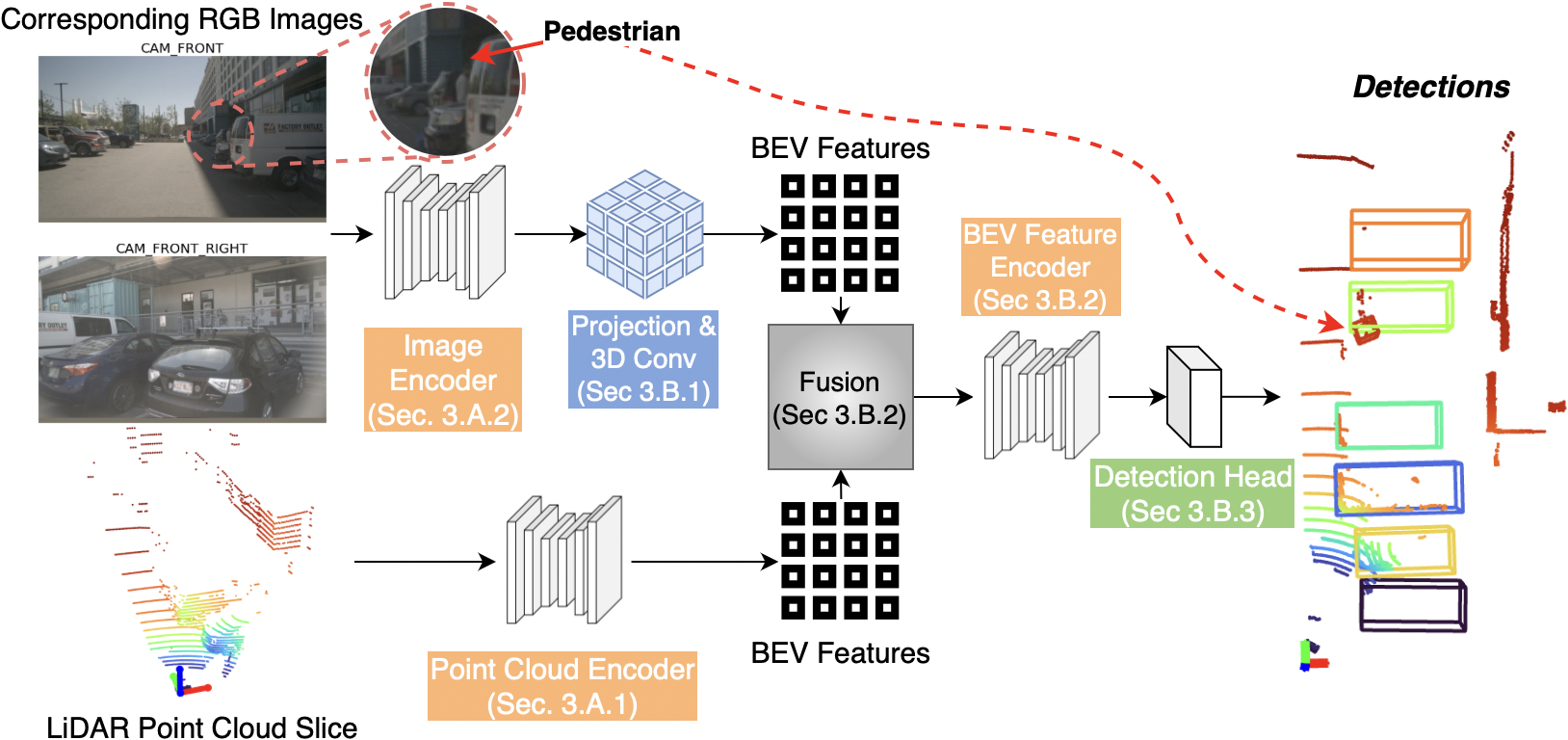}
  \vspace{-0.2cm}.
  \caption{Overview of the proposed framework. With a point cloud slice and its corresponding images as the input (see cars and pedestrian), two modalities are encoded separately and in parallel, producing two top-down BEV feature maps. These maps are fused and sent to a BEV encoder and a detection head. The example shows the importance of complimentary and spatially-related feature maps: Images provide the context needed to accurately detect the fragmented bottom vehicle with very few points, and the person is detected despite occlusion and shade with the support of the point cloud input.}
  \label{fig:arch}
  \vspace{-0.5cm}
\end{figure*}

Meanwhile, cameras are widely deployed on AVs and usually acquire images at a high rate. Images also provide a wider FoV than a LiDAR slice, which makes them ideal in providing a global context for streaming detection. Dominant camera-LiDAR fusion methods \cite{contfuse, epnet, pointpainting, wang2021pointaugmenting, percaware} rely on point-wise operations and require a matching FoV between LiDAR and camera data. In our streaming setting however, the FoV overlap between a camera and a narrow LiDAR slice is small. Thus, existing fusion methods are not suited for our problem.



To bridge the above research gap, we propose a multi-modal streaming 3D object detection framework that utilizes camera images to provide a \emph{current} global context to a LiDAR slice using an innovative fusion method. The difference in FoV between the two modalities is addressed by extracting the features of each modality (an image and a LiDAR slice) independently and in parallel. These features are then projected to a unified spatial Birds Eye View (BEV) representation. This representation is processed by convolutional neural networks (CNNs) and sent to a detection head for 3D bounding box prediction. We also propose a feature map cropping operation that significantly speeds up the forward pass while maintaining cross-modal spatial relations. Our contributions are as follows:


\begin{enumerate}
    \item We propose the first framework for multi-modal streaming detection that effectively uses camera images to complement narrow LiDAR slices.
    \item This framework outperforms prior streaming methods without relying on past LiDAR slices. It also outperforms representative single and multi-modal full-scan detectors while taking only a LiDAR slice as input.
    \item Our method encodes each modality separately and does not depend on past slices. This parallelism can be exploited by GPUs or FPGAs to further reduce latency.
    \item Experiments show the benefits of the proposed method and its robustness to small camera-LiDAR miscalibration, missing camera images, and very narrow LiDAR slices (see Fig. \ref{fig:delay}). We also discuss the runtime and computational cost of the proposed framework.
\end{enumerate}

\section{RELATED WORK}
\label{sec:rel}


\paragraph{LiDAR-Based 3D Detection}
Processing a LiDAR point cloud for 3D object detection usually involves one of the following three representations: The 1st is voxelizing the input into 3D grids and extracting features using 3D CNNs or sparse convolutions \cite{voxelnet, second}. While accurate, these models are computationally expensive and depend on voxel resolution. The 2nd is processing the point cloud as a dense range image \cite{lasernet, fan2021rangedet}. This representation, while efficient, suffers from object overlap, and sizes varying with depth. The third and most popular is the top-down BEV which maintains constant object sizes, although it is much sparser than range images \cite{centerpoint, pointpillars}. This representation has shown great performance for 3D detection since it simplifies the localization task and utilizes 2D CNNs for efficiency. 

\paragraph{LiDAR-Camera Fusion}
Early fusion work relied on point-wise sampling of image features across multiple scales \cite{contfuse, epnet}. However, these methods are complicated and often suffer from feature blurring. Recently, simpler and more effective methods proposed decorating each 3D point with image features \cite{wang2021pointaugmenting} or segmentation scores \cite{pointpainting}. Most prior work on fusion combines information from both modalities in a point-wise fashion and requires matching FoVs between the two modalities. Point-wise methods are thus not suited for our streaming problem where the camera images provide a wider FoV than the narrow LiDAR slices.

Moreover, point-wise fusion methods require extremely accurate camera-LiDAR calibration which can degrade with operation \cite{cui2021deep}. Yoo \textit{et al.} \cite{3dcvf} use auto-calibration to project image features to BEV and attention mechanisms to fuse multi-modal features during encoding and at the region refinement stage. Our method differs from prior work in that it is fully convolutional, designed specifically for streaming detection, avoids strict point-wise fusion, and significantly improves the LiDAR-only baseline.

\paragraph{Streaming 3D Object Detection}
\label{sec:stream}
Streaming detection is a relatively new direction with limited prior work. Nonetheless, it was shown to be effective in replacing full $360^\circ$ sweeps for more latency-efficient perception. Han \textit{et al.} \cite{stream} proposed modifying the meta-architecture of full-scan 3D detectors to facilitate operating on streaming LiDAR slices. They used recurrent neural networks to preserve features from past slices and pass them to the current slice, thus expanding the context and restoring the degraded accuracy due to streaming. 
Frossard \textit{et al.} \cite{strobe} proposed storing multi-scale feature maps from past slices and concatenating them with current slice features. Finally, Chen \textit{et al.} \cite{polarstream} pointed that slices are better represented in polar coordinates. They used BEV ``polar pillars'' and padded past slice features to current features along the azimuth dimension.

Such past-aware solutions were able to compensate for most of the context lost due to the narrow streaming input. However, we argue that such good results are mainly a consequence of an unrealistic experimental setup. All prior streaming detection works simulate a streaming input by slicing a relatively \emph{static} LiDAR scan (collected by slow vehicles) into $n$ slices. This makes the same false assumption (that LiDAR points are observed synchronously) as full-scan models; and therefore past features from \emph{many} prior slices are sufficient to replenish the model. However in reality, a streaming input is dynamic and past features would represent an outdated state of the world as discussed in Sec. \ref{sec:intro}. This is a main motivation behind our work: we replace past LiDAR slices with up-to-date images.

\section{Multi-modal Streaming 3D Detection}

\label{sec:method}
An overview of the proposed framework is shown in Fig. \ref{fig:arch}. It takes as input a LiDAR point cloud slice and camera images encompassing the slice and outputs detected objects within a slice. Each modality goes through a separate encoding pipeline for feature extraction (Sec. \ref{sec:methdod/feats}). We present a novel method to project dense 2D image features to BEV using an intermediate 3D volumetric representation. The resulting multi-modal BEV representations are then fused, encoded, and sent through a detection head (Sec. \ref{sec:methdod/fusion}). Finally, to speed up inference, a feature map cropping method is proposed in Sec. \ref{sec:fold}.

\subsection{Multi-modal Feature Extraction}
\label{sec:methdod/feats}

Features from both input modalities are extracted independently and in parallel. The goal is to have these features in a common spatial (BEV) and feature space for fusion. 

\subsubsection{Point Cloud Encoding} First, the 3D point cloud slice is transformed into a BEV feature map. A PointPillars \cite{pointpillars} encoder is adopted as it well-balances accuracy and speed. Each point is represented with a vector $(x,y,z,r,m,s)$ where $x,y,z$ denote the 3D location, and $r,m$ denote reflectance and relative timestamp. We add the slice index $s$ to allow the model to learn slice-specific information. The scene is divided into a top-down BEV $V_x\times V_y$ grid. Points within each pillar (grid pixel) are encoded using a PointNet \cite{qi2017pointnet}. This gives each pillar a feature vector of size $c$. Overall, this stream produces a 2D BEV feature map $\mathbf{P}_{bev}\in\mathbb{R}^{V_x\times V_y \times c}$.

\subsubsection{Camera Image Encoding}
Let $\mathbf{X}_j \in \mathbb{R}^{H\times W \times 3}$ be the $j$-th input image from $J$ images. A pre-trained ResNet-50 \cite{resnet} backbone is used to extract multi-scale 2D feature maps from each image. Afterwards a feature pyramid network (FPN) \cite{fpn} is used to aggregate these feature maps into $\mathbf{F}_j \in \mathbb{R}^{\frac{H}{4}\times \frac{W}{4} \times c}$. This stream produces $J$ semantically-rich 2D feature maps from $J$ images around a LiDAR slice.

\subsubsection{Discussion}
Feature extraction for each modality is separate. This allows the fusion process to be independent of point-wise or sequential operations. This is motivated by three reasons: First, independent encoding allows for retaining the complementary information from both modalities. For example, image features are not simply appended to sparse and narrow point cloud slices \cite{pointpainting} and thus losing their richness and density. This allows image features to provide needed context that is wider than the slice. Second, independent encoding boosts robustness to small camera-LiDAR miscalibration, missing points, or missing camera images as discussed in Sec. \ref{sec:exp}. Third, parallel encoding allows for better utilization of parallelization hardware.

\begin{figure}[t]
\centering
\includegraphics[width=\linewidth]{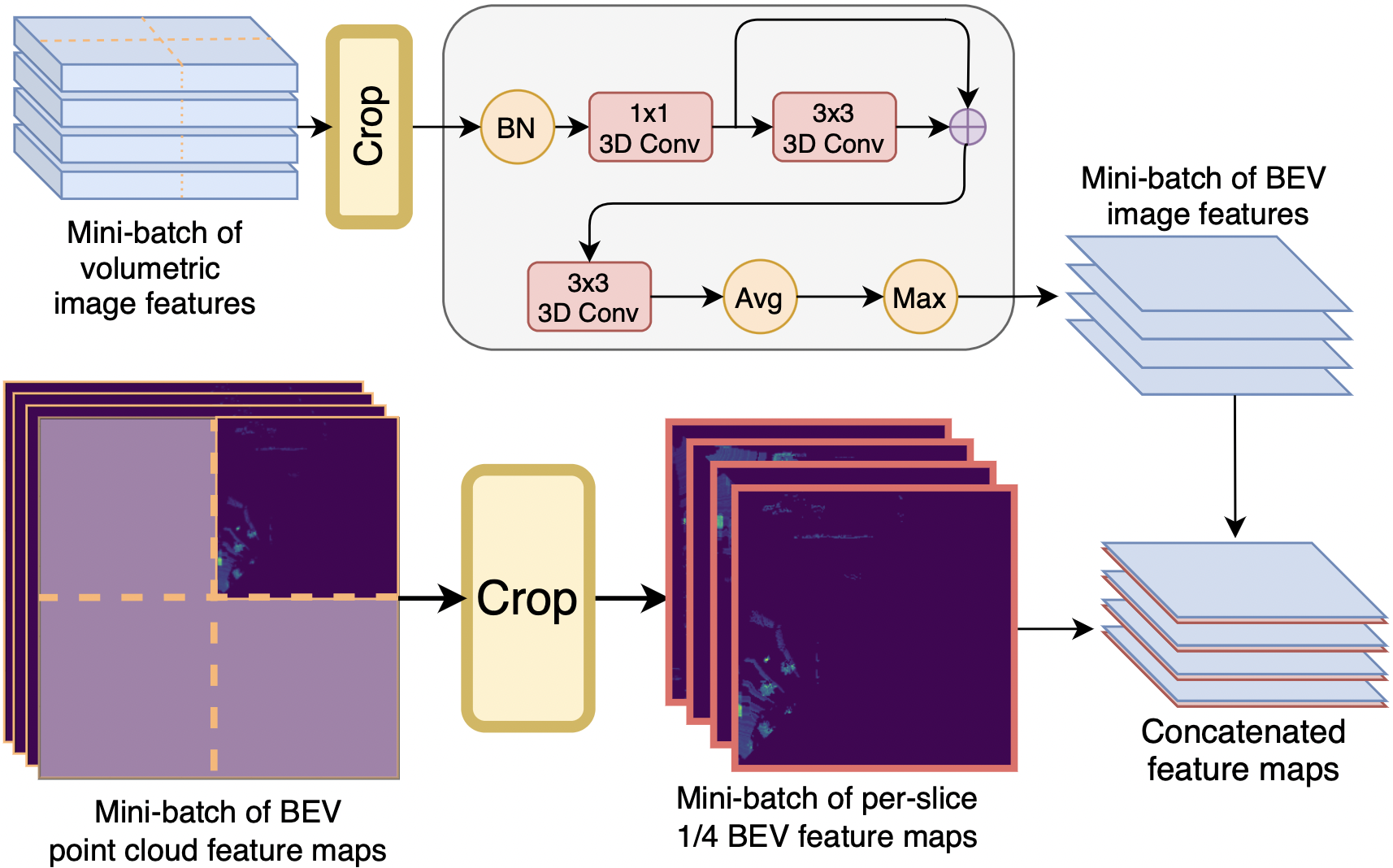}
\caption{Top: Transforming volumetric image features to BEV. Bottom: Divide a feature map/volume into 4 patches, discard empty regions, and stack patches that contain slice features in a mini-batch for training. A similar procedure is used to crop the 3D volumetric representation which significantly reduces compute. Finally, cropped multi-modal BEV feature maps are concatenated. BN: Batch norm. Avg: average pooling along the z-axis. Max: max pooling to eliminate the z-axis.}
\label{fig:fusion}
\vspace{-0.3cm}
\end{figure}
\subsection{Multi-modal Fusion}
\label{sec:methdod/fusion}



\subsubsection{Projection and 3D Convolutions} 
To widen the 3D context around a LiDAR slice, 2D image features must be projected to a common spatial and feature space with the point cloud features. Given $J$ image feature maps $\mathbf{F}_j$, projection is done as follows: The scene is divided into a 3D grid $\mathbf{V}\in\mathbb{R}^{V_x\times V_y\times V_z \times c}$. This is the same voxel size used to grid the point cloud input in the $xy$ plane. However, instead of pillars there are multiple $0.5m$-high voxels along the z-axis. Projection is done using perspective mapping $\mathbf{\Pi}$ from the pinhole camera model, and the intrinsic $\mathbf{K}$ and extrinsic $\mathbf{R}_t$ matrices of each camera. To project an $\mathbf{F}_j$ pixel $(u,v)$ to 3D voxels $(x,y,z) \in \mathbf{V}$ we use Eqn. \ref{eq:proj}. All voxels along a ray that passes through a $F_j$ pixel will be filled with its features. Voxels outside the FoV of a camera are padded with zeros. If multiple pixels project to the same 3D voxel (intersecting camera FoVs), their features are averaged. Now image features are in 3D space $\mathbf{V}$.
\begin{align}
  [u,v]^T =\mathbf{\Pi} \mathbf{KR}_t [x,y,z,1]^T
  \label{eq:proj}
\end{align}

The projected image features are then transformed from the 3D volumetric representation $\mathbf{V}$ to a BEV representation $\mathbf{I}_{bev} \in\mathbb{R}^{V_x\times V_y \times c}$ using a few modules shown in Fig. \ref{fig:fusion}. First, batch norm is used to share statistics between mini-batches and stabilize the output from the image FPN. This benefits from the cropping and stacking operation discussed in Sec. \ref{sec:fold}. To bridge the gap between the feature spaces of our two modalities, we use 1x1 convolutions as a projection layer from the image feature space to the point cloud feature space. Non-linear activation (ReLU) is also used to model complex projections in high dimensional space.

This is followed by a residual module consisting of a single 3D convolution with kernel size 3 followed by batch norm and ReLU. Afterwards, another 3D convolution (plus batch norm and ReLU) is used with stride 2 in the z-axis. This effectively halves the number of z-axis voxels and gives them a height of $1m$. Then, average pooling with stride 2 in the z-axis is used to obtain $2m$-high voxels. Two meters usually cover the height of common objects of interest (i.e. cars, or pedestrians). Finally, the max operator is used along the $z$ dimension to reduce it to 1, effectively obtaining a BEV feature map $\mathbf{I}_{bev}$. We use very few 3D convolutions and rely on pooling to mitigate the computational cost. 

\subsubsection{Fusion}
The resulting image $\mathbf{I}_{bev}$ and point cloud $\mathbf{P}_{bev}$ representations are spatially-related in BEV. Fusion is then a concatenation of the BEV feature maps channel-wise. This contextually-rich combined 2D representation $\mathbf{S}_{bev} \in \mathbb{R}^{V_x\times V_y \times c*2}$ is then sent through a lightweight CNN-based BEV encoder. This encoder outputs 3 multi-scale feature maps which are then aggregated with an FPN \cite{fpn} and sent to the detection head.

\subsubsection{Detection Head}
\label{sec:methdod/head}
We use a CenterPoint head \cite{centerpoint}, which is an anchor-free method to localize and regress 3D bounding boxes. This head utilizes heatmaps for object center regression, bounding box estimation and orientation prediction. While the original implementation uses separate heads each focused on classes of a common size, we follow \cite{polarstream} and use a single head for all classes. This significantly speeds up inference as discussed in Sec. \ref{res:abl}. It has also proven sufficient when coupled with per-class non-maximum suppression (NMS) for post-processing.

\subsection{Crop and Uncrop}
\label{sec:fold}

As shown in Fig. \ref{fig:fusion}, during the forward pass, most of the BEV feature map is empty. That is because narrow LiDAR slices do not take up much physical space. Similarly, when one or two cameras are used (total FoV $\sim 125^\circ$), most of the 3D volumetric representation $V$ is empty after projection. Convolutions on these empty regions wastes memory and compute. To save these resources and to facilitate convergence, we propose dividing a single BEV feature map into four quarters, discard the ones without a slice and stack the per-slice \nicefrac{1}{4} feature maps to process them as a mini-batch. The same cropping procedure is used with the volumetric representation ($V_{cr}\in\mathbb{R}^{\nicefrac{V_x}{2}\times \nicefrac{V_y}{2}\times V_z\times c}$). This maintains the spatial correspondence between BEV feature maps from both modalities. It also speeds up projection and 3D convolutions.

Different slice feature maps within a batch are likely to share some information. This can be exploited using the batch norm operation \cite{batchnorm} which has been shown to greatly stabilize and speed up training convergence. Cropping also reduces the size of the input space to our BEV encoder. Instead of learning to infer from the whole scene, the model needs to attend only to the smaller space around the slice. This can make the detection task more tractable. 

During inference, cropping significantly reduces latency, especially for the 3D convolutions used in projection. Results on speed and FLOPs are reported empirically and discussed further in Sec. \ref{res:abl}. After fusion and processing of all \nicefrac{1}{4} BEV feature maps through the BEV encoder, the feature maps are "uncropped'' back into the original BEV map size by zero-padding. This restores the overall BEV map before processing by the detection head.
\setlength{\tabcolsep}{5pt}
\begin{table}[t]
    \begin{center}
        \caption{NuScenes test set: 3D detection accuracy results}
        \label{table:testset}
        \resizebox{\linewidth}{!}{
        \begin{tabular}{l|c|c|c|ccc}
            \hline\noalign{\smallskip}
            Model & Modality & Stream & mAP & Car & Ped & Bicycle  \\
            \noalign{\smallskip}
            \hline
            \noalign{\smallskip}
            PointPillars \cite{pointpillars}  & PC & \xmark & 30.5 & 68.4 & 59.7 & 1.1 \\
            PointPainting \cite{pointpainting} & PC+RGB & \xmark & 46.4 & 77.9 & 73.3 & 24.1 \\
            3D-CVF \cite{3dcvf} & PC+RGB & \xmark & 52.7 & 83.0 & 74.2 & 30.4 \\
            CenterPoint \cite{centerpoint}  & PC & \xmark & 60.3 & 85.2 & 84.6 & 30.7 \\
            \noalign{\smallskip}
            \hline
            \noalign{\smallskip}
            PolarStream \cite{polarstream} (4s) & PC & \cmark & 52.9 
            & 80.7
            & 78.1
            & 29.6 \\
            Ours (8s) & PC+RGB & \cmark & \textbf{53.8}
            & \textbf{81.2}
            & \textbf{80.5}
            & \textbf{42.5} \\
            \hline
        \end{tabular}}
    \end{center}
    \vspace{-0.5cm}
\end{table}
\setlength{\tabcolsep}{1.4pt}

\section{Experimental Setup}
\label{sec:exp}

\subsection{Streaming LiDAR Input}
\label{sec:method/stream}
Most available LiDAR datasets provide full $360^\circ$ scans. To simulate a streaming LiDAR input, a full LiDAR sweep is divided into $n$ non-overlapping slices (with an angle of $\frac{360^\circ}{n}$). This slicing method follows prior work, however, our framework does not make any assumptions about past slices since it does not use them. For training, each slice is annotated with ground truth bounding boxes if a single corner or more are within the slice sector. This trains the network to adapt to cases where an object is heavily fragmented or even without points in a slice. The predefined FoVs of cameras are used to determine if a camera overlaps with a slice.

\subsection{Dataset and Implementation Details}

We evaluate on the popular large-scale NuScenes \cite{nuscenes} dataset because it provides calibrated and synchronized $360^\circ$ LiDAR scans and camera images, which is essential for our problem of interest. The dataset contains 700 sequences (each 20s long) for training, 150 for validation and another 150 for testing. 
The dataset is built using a 32-beam LiDAR and 6 cameras with intersecting FoVs that cover all sides of an AV, each producing an image with a resolution of $1600\times900$. Each camera covers an angle of $70^\circ$ except the back camera which covers $110^\circ$. It also provides transformation matrices for LiDAR-to-Camera projection and vice versa. 

The ResNet backbone is obtained from a hybrid task cascade model \cite{hybrid} that is pre-trained on 2D detection and segmentation for autonomous driving scenes \cite{nuscenes}. 

The scene's grid range is set to $[-51.2, 51.2]m$ for the $x$ and $y$ axes, and $[-3, 5]m$ for the $z$ axis. For the point cloud stream, the voxel size is set to $0.2\times0.2\times8$, producing a $512\times512$ feature map in BEV. For the volumetric representation of image features, the voxel size is set to $0.2\times0.2\times0.5$, producing a $512\times512\times16$ volumetric representation for image features. After 3D convolutions, we obtain a BEV feature map of shape $512\times512$ matching the point cloud BEV feature map. After cropping, the feature map is $256\times256$ for both modalities.

For evaluation, the dataset uses the mean average precision (mAP) metric based on the accuracy of center detection within 4 thresholds: \{0.5, 1, 2, 4\}$m$. The overall mAP metric is the average over classes and center thresholds.

Of the limited prior work on streaming detection, \cite{stream} used a public LiDAR dataset which does not provide $360^\circ$ camera images, \cite{strobe} used an in-house dataset, and \cite{polarstream} used NuScenes and re-trained prior work \cite{stream, strobe} on it. Thus, we evaluate and compare with prior work on the NuScenes benchmark.

\setlength{\tabcolsep}{6pt}
\begin{table}[t]
    \begin{center}
        \caption{NuScenes val set: mAP results of streaming models when using different no. of slices}
        \label{table:stream}
        \resizebox{\linewidth}{!}{
        \begin{tabular}{l|c|c|ccccc}
            \hline\noalign{\smallskip}
            Model & Modality & Past Features & 1 & 4 & 8 & 16 & 32\\
            \noalign{\smallskip}
            \hline
            \noalign{\smallskip}
            Han et al. \cite{stream}  & PC & \cmark & - & 52.9 & 53.8 & 52.7 & 50.6\\
            Strobe \cite{strobe} & PC & \cmark & 46.9 & 49.4 & 47.9 & 45.4 & 42.0\\
            PolarStream \cite{polarstream} & PC & \cmark & - & 53.2 & 52.7 & \textbf{53.9} & \textbf{52.4}\\
            \noalign{\smallskip}
            \hline
            \noalign{\smallskip}
            CenterPoint$^{\mathrm{p}}$ \cite{centerpoint} & PC & \xmark & 49.1 & 30.4 & 27.1 & 22.6 & 2.2\\
            Ours & PC+RGB & \xmark & \textbf{57.6} 
            & \textbf{55.3} 
            & \textbf{54.7} 
            & 53.8 
            & 51.1 
            \\
            \hline
        \end{tabular}}
    \end{center}
\end{table}

\subsection{Training and Inference Details}
A pre-trained ResNet-50 backbone is used only for obtaining image features. No data augmentation is used due to the cross-modal nature of our work. However, we use the class-balanced sampling protocol proposed in \cite{cbgs} to address the class imbalance of NuScenes. At inference, the top 500 detections are gathered and then per-class NMS is performed within a slice. The dataset we use is not made for streaming detection or per-slice evaluation. For a fair comparison with other methods, detections are aggregated from previous slices in a scene as a post-processing step and then per-class NMS is performed to filter out duplicated detections between slices \cite{stream}. In general, NMS cannot help if a model does not produce accurate detections and post-processing depends on downstream tasks such as prediction.

\emph{\textbf{Note on training speed:}} After multiple rounds of hyper-parameter tuning and using different optimization schemes, it was found that our model usually yields good results after 1 or 2 epochs and then it starts to overfit. This contrasts greatly with LiDAR-only models which usually require training for up to 20 epochs. We hypothesize that the dense and semantically-rich image features provide strong priors to the model. Moreover, the cropping operation limits the input space the model has to learn from. It also allows for sharing statistics between slices with batch norm \cite{batchnorm}, which has been shown to greatly facilitate convergence.

\vspace{0.1cm}
\section{Results and Discussion}
\begin{figure*}[t]
\centering
\includegraphics[width=\textwidth]{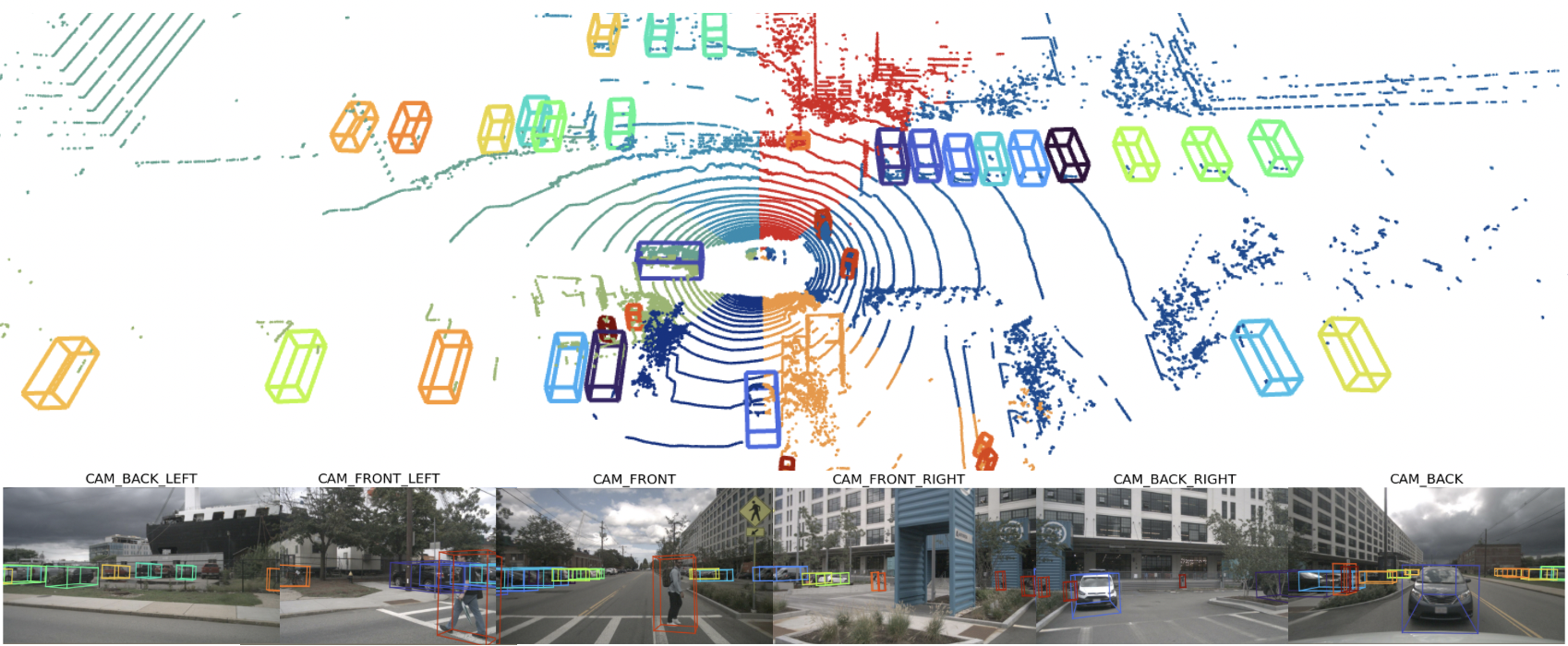}
\caption{Qualitative example of our streaming model's performance on the NuScenes validation set. Detections from 8 slices are aggregated (each slice is colored) and NMS is used to filter duplicates. Though the cars in the `back' and `back right' cameras (as well as other objects) are fragmented between slices, they are still accurately detected.}
\label{fig:example}
\end{figure*}

We compare the proposed model with both streaming and non-streaming models. These include LiDAR-only, camera-only, and fusion models. Moreover, the robustness of our model to missing camera images and camera-LiDAR miscalibration is studied and discussed. We also report ablation studies related to runtime, model design, and the effect of different image resolutions. Qualitative examples are demonstrated in Fig. \ref{fig:back} and \ref{fig:example}.

\subsection{Detection Benchmark Results}
Evaluated on the NuScenes test set, our \emph{8-slice} model achieves the state-of-the-art (SOTA) mAP for streaming models without using any past features (see Table \ref{table:testset}). This accuracy surpasses the 4-slice PolarStream \cite{polarstream} model, showing that even with more fragmentation and without access to past information, our model performs better for detection. For the test set, our model outperforms popular full-sweep detectors and baselines (both LiDAR-only and fusion-based ones). It is worthy noting that our results were obtained without using data augmentation in order to maintain cross-modal consistency. As shown in prior work \cite{second, centerpoint, wang2021pointaugmenting}, augmentation can significantly boost mAP.


\setlength{\tabcolsep}{6pt}
\begin{table}[t]
    \begin{center}
        \caption{NuScenes val set: comparison with non-streaming models}
        \label{table:nonstreamcompare}
        
        \resizebox{\linewidth}{!}{
        \begin{tabular}{@{}l|c|c|cccc}
            \hline\noalign{\smallskip}
            \textbf{Model} & \textbf{Modality} & \textbf{Stream} & \textbf{mAP} & \textbf{Car} & \textbf{Ped} & \textbf{Bicycle}\\
            \noalign{\smallskip}
            \hline
            \noalign{\smallskip}
            PointPillars \cite{pointpillars} & PC & \xmark & 29.5 & 70.5 & 59.9 & 1.60\\
            ImVoxelNet \cite{imvoxelnet}  & RGB & \xmark & - & 51.8 & - & -\\
            3D-CVF \cite{3dcvf} & PC+RGB & \xmark & 42.2 & 79.7 & 71.3 & -\\
            MoCa \cite{zhang2020exploring} & PC+RGB & \xmark & 47.9 & 82.4 & 79.1 & 27.2 \\
            CenterPoint$^{\mathrm{p}}$ \cite{centerpoint} & PC & \xmark & 49.1 & 83.8 & 77.4 & 12.7\\
            Ours (all) & PC+RGB & \xmark & \textbf{57.6}  & \textbf{83.8} & \textbf{80.9} & \textbf{51.6} \\
            \noalign{\smallskip}
            \hline
            \noalign{\smallskip}
            CenterPoint$^{\mathrm{p}}$ (8s)  & PC & \cmark & 27.1  & 66.1  & 59.7  & 4.9 \\
            Ours (8s) & PC+RGB & \cmark & 54.7  & 83.1  & 81.8 & 47.3 \\
            Ours (16s) & PC+RGB & \cmark & 53.8  & 82.5  & 81.2 & 52.1 \\
            Ours (32s) & PC+RGB & \cmark & 51.1  & 80.6  & 79.0  & 47.3\\
            \noalign{\smallskip}
            \hline
            Ours (8s)$\dagger$ & PC+RGB & \cmark & 47.8  & 80.4  & 76.9 & 38.0 \\
            \hline
            \multicolumn{7}{l}{$^{\mathrm{p}}$Pillar-based. $\dagger$without the back camera image. } \\
        \end{tabular}}
    \end{center}
\end{table}
\setlength{\tabcolsep}{1.4pt}

\subsubsection{Comparisons with Streaming Models}
As shown in Table \ref{table:stream}, our streaming mAP is on par or exceeds prior streaming methods without using memory mechanisms or features from past LiDAR slices. Current image features are able to effectively replace past LiDAR features and provide sufficient context for accurate streaming detection. Moreover, our 8-slice model performs better than models that take the whole scene or wider slices (4) as input. We believe 8-slice is generally a good middle ground between too few or too many slices. It is enough to reduce latency and still contains enough contextual information. As a baseline, we implement a streaming variant of a comparable and powerful LiDAR-only detector (Pillar-based CenterPoint$^{\mathrm{p}}$ \cite{centerpoint}) by limiting its input to one slice during training and evaluation. As shown in Table \ref{table:stream}, our RGB representation was able to restore and surpass all lost accuracy due to streaming, when compared to the streaming CenterPoint$^{\mathrm{p}}$. Even at detrimental heavy fragmentation (32 slices) our model increases mAP from 2.2 to 51.1. This 32-slice mAP exceeds that of the full-sweep CenterPoint$^{\mathrm{p}}$ shown in Table \ref{table:nonstreamcompare}. Also, our full-scan model was able to increase mAP for CenterPoint$^{\mathrm{p}}$ by 8.5, showing the benefit of our fusion mechanism even for full-scan models. Finally, maintaining mAP $>50$ between all slicing scenarios shows that the image features provide valuable and stable context that can withstand point cloud fragmentation (see Fig. \ref{fig:delay}). Results of prior streaming models on NuScenes validation set were reported in \cite{polarstream}.

\begin{figure}[t]
\centering
\includegraphics[width=0.8\linewidth]{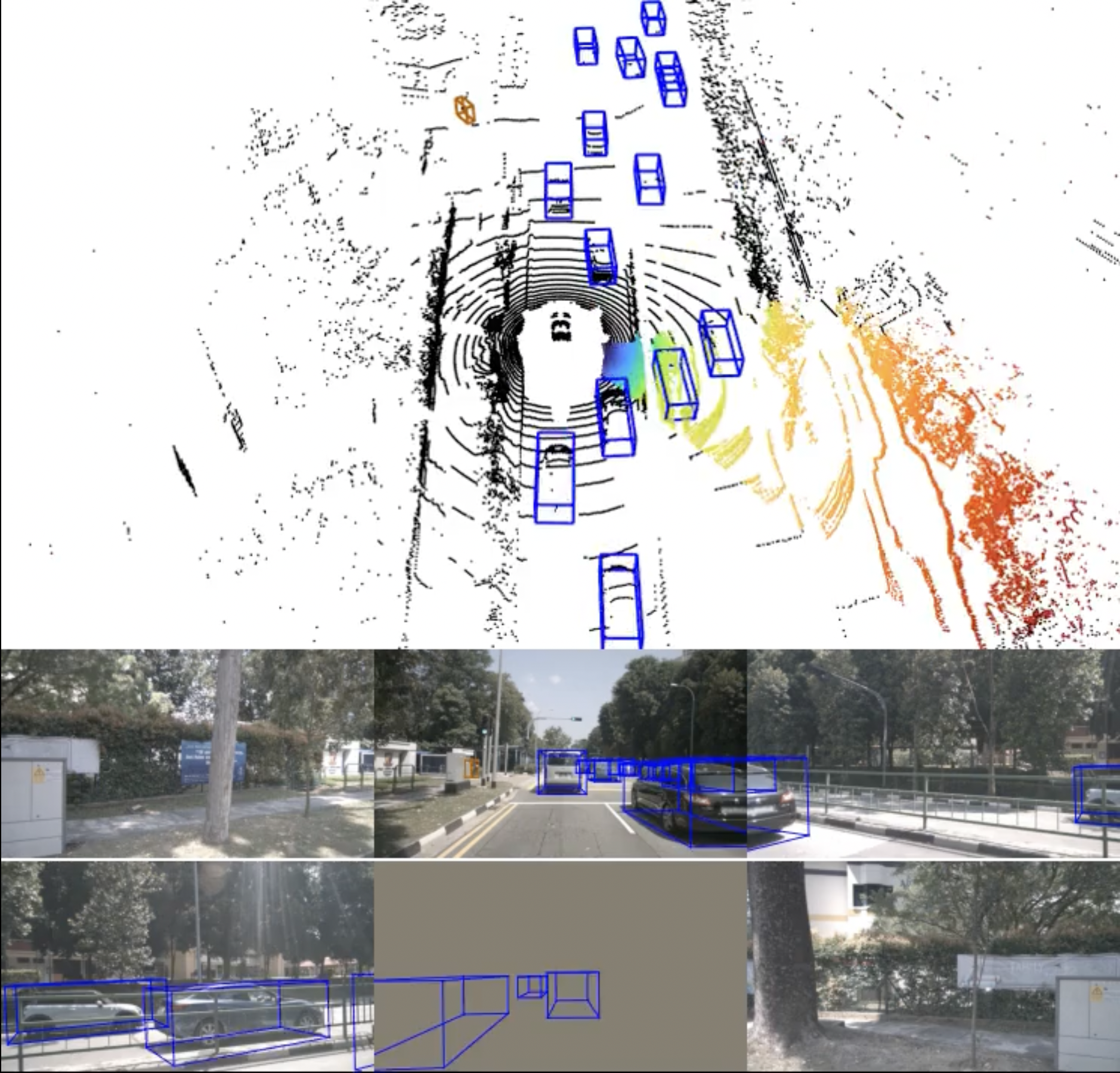}
\caption{Qualitative example of the performance of the 8-slice model without the back camera input. The highlighted point cloud shows the last LiDAR slice input. Detections from all slices are aggregated for visualization.}
\label{fig:back}
\vspace{-0.2cm}
\end{figure}

\subsubsection{Comparisons with Non-streaming Models}
Taking as input a fraction of the 3D scene, our model is able to detect objects even when heavily fragmented (for 8, 16, and 32 slices) with an accuracy higher than prior full-sweep models on 3D detection and a similar LiDAR-only detector (CenterPoint$^{\mathrm{p}}$). As shown in Table \ref{table:nonstreamcompare}, without access to past information, mAP degrades significantly for the streaming CenterPoint. However, our proposed fusion method provides enough context to replenish an impoverished point cloud input. Most notable, our streaming models are able to improve the accuracy for small object detection, especially for the bicycle class. This is likely because: a) bicycles have a small point cloud footprint with inconsistent shapes compared to other objects like cars and pedestrians; and b) bicycle is a tail class on NuScenes \cite{cbgs}. For both reasons, it benefits greatly from semantically-rich image features.

\subsubsection{Robustness to missing images/few points}
We experiment with training without the image from the back camera (which has a wider FoV than other cameras) and note that our mAP degrades only slightly for the critical car and pedestrian classes (the last row of Table \ref{table:nonstreamcompare}). This shows that our model is somewhat robust to missing camera images. This is shown qualitatively as well by visualizing accurate detections behind the ego-vehicle in Fig. \ref{fig:back}. Moreover, at heavy fragmentation and with a limited number of points, our 32-slice model maintains high accuracy. These observations illustrate the benefit of decoupling the image and point streams for a robust complimentary multi-modal representation. 

\subsection{On Calibration}
Point-wise fusion models depend heavily on accurate camera-LiDAR calibration, which can be difficult to maintain during operation due to mechanical vibrations or thermal strains \cite{cui2021deep, rggnet}. This dependence comes from decorating points directly with image features or segmentation scores \cite{pointpainting, wang2021pointaugmenting, mvxnet}. In contrast, our model's decoupling of point/image encoding and its relatively sparse volumetric representation for image features lead to better robustness to small calibration errors. To study the effect of calibration error, we survey SOTA work on online self-calibration (traditional \cite{taylor} and learning-based \cite{rggnet, lccnet, semalign}) and find that they can bring the mean Euler angle error down to around $\pm1^\circ$ and mean translation error to $\pm10 cm$ from the ground truth calibration depending on the level of noise.

\begin{figure}[t]
\centering
\includegraphics[width=\linewidth]{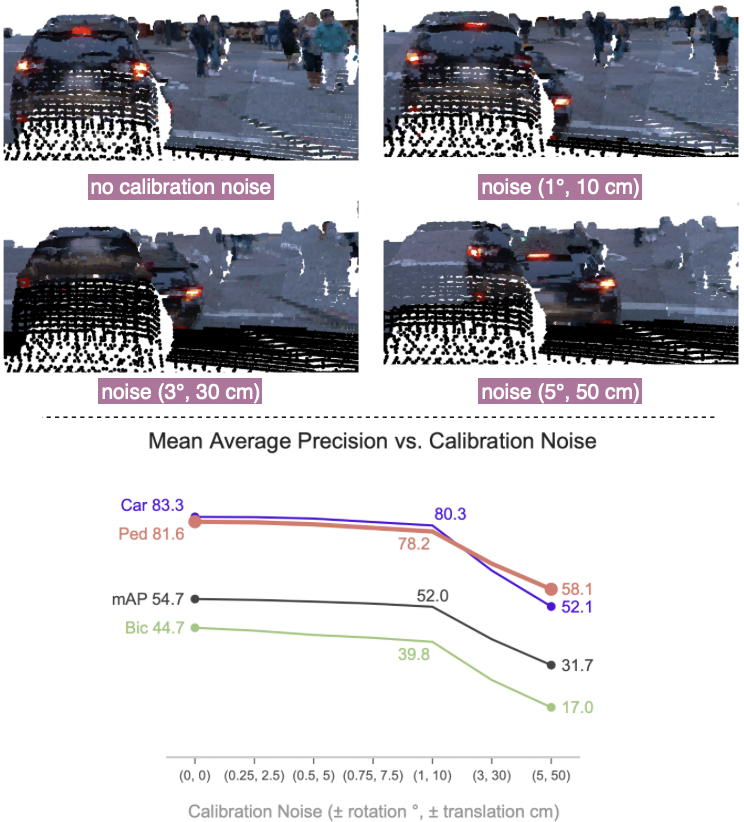}
\caption{Top: Projection of RGB values to 3D points using noisy extrinsic matrices with various degrees of noise. Even small noise completely disappears some pedestrians and corrupts vehicle points posing a critical danger to point-wise fusion methods. Bottom: Effect of the calibration noise on the mAP of our 8-slice model.}
\label{fig:calib}
\end{figure}

To simulate the error, we follow prior work in self-calibration by uniformly sampling (within a $\pm$ range) a translation vector and Euler angles (producing a noisy rotation matrix). These noisy transformations are applied to the ground truth extrinsic matrix $R_t$. We test on 4 thresholds within the expected error and on 2 worst case scenario errors at $(3^\circ, 30 cm)$ and $(5^\circ, 50 cm)$. As shown in Fig. \ref{fig:calib}, even the $(1^\circ, 10 cm)$ calibration error can greatly change the pixel-point correspondence where small and critical objects like pedestrians are almost completely covered by road features. Such issues are exacerbated with distance. The work in \cite{wang2021pointaugmenting} has shown that even with accurate calibration, false negatives in segmentation masks can be detrimental for models that use point-wise decoration with segmentation scores \cite{pointpainting}. Point-wise features projected with $(1^\circ, 10 cm)$ noise would be deeply problematic. In contrast, our model shows robustness by maintaining strong performance at that noise level with only a 2.7 degradation in mAP. However, at very high noises, accuracy degrades indicating that image features do provide valuable context for streaming multi-modal detection. At all noise levels, our mAP performance remains higher than the full-scan LiDAR-only baseline \cite{pointpillars}.

\setlength{\tabcolsep}{3pt}
\begin{table}[t]
    \begin{center}
        \caption{FLOPs and Runtime for various components for the 8-slice model. We benchmark on one image and 30000 points as input.}
        \label{table:timing}
        \resizebox{\linewidth}{!}{
        \begin{tabular}{c|ccc|c}
            \hline\noalign{\smallskip}
             Module & GFLOPS & Runtime (ms) & Runtime/fp16 (ms) & mAP\\
            \noalign{\smallskip}
            \hline
            \noalign{\smallskip}
            Pts. + BEV Enc. + Multi-Head & 180.2 & 44.5 & 33.8 & 54.7\\
            Pts. + BEV Enc. + Single-Head & 161.6 & 27.4 & 18.2 & 54.7\\
            Above w/ cropping & \textbf{40.43} & \textbf{15.8} & \textbf{14.0} & 54.7 \\
            3D Convolutions & 715.7 & 157.6 & 61.6 & -\\
            Above w/ cropping & \textbf{178.9} & \textbf{38.5} & \textbf{15.9} & -\\
            \noalign{\smallskip}
            \hline
            \noalign{\smallskip}
            Img Encoding (1600x900) & 126.5 & 28.9 & 14.8 & 54.7\\
            Above w/ (1200x675) & 71.6 & 17.2 & 11.7 & 53.9\\
            Above w/ (800x450)  & \textbf{32.1} & \textbf{9.8} & \textbf{11.7} & 50.8\\
            \hline
        \end{tabular}}
    \end{center}
    \vspace{-0.4cm}
\end{table}
\subsection{On Parallelization}
\label{res:abl}

While prior streaming work can be more light-weight, our work facilitates faster runtimes by allowing better parallelization. This is accomplished by removing multiple sequential dependencies. First, in the proposed framework, point encoding does not depend on image features. Thus, encoding of both modalities can run in parallel. Second, our framework does not require past slice features which allows for processing slices in parallel as they arrive.

Considering the parallelism inherent in the proposed method and the fast frame rate of cameras, our streaming framework can be efficient in terms of the end-to-end runtime. For example, an AV can take advantage of this parallelism by maintaining our proposed 3D volumetric representation of the world and simply updating it as images arrive from cameras. In that case, the LiDAR-only stream only has to fetch the most up-to-date features from the image's BEV representation. Thus the end-to-end runtime (starting from a point cloud slice in the input, and ending with detections in the output) of our 8-slice streaming model is only \emph{15.8 ms} with SOTA accuracy. Also, since our framework is fully convolutional, it benefits from the advances in neural network quantization and pruning \cite{martinez2021permute}. Finally, this parallel streaming approach allows the computational burden of perception to be distributed along the scan instead of processing the whole scene at once, or in a sequence.

We study runtime and FLOPs of various components of our (8-slice) model on a Tesla V100 GPU and report the results in Table \ref{table:timing}. A slice ($45^\circ$) can be covered by a single image based on the FoV of Nuscenes cameras ($70^\circ$) and contains about 30000 points. These values are used during runtime testing. Moreover, runtime is reported at full and half precision (fp16) since many latency-critical applications use fp16 to speed up inference and reduce storage requirements. Table \ref{table:timing} shows the significant impact of our cropping operation on 3D convolutions ($\sim$75\% reduction in runtime and FLOPs) and on our LiDAR-only pipeline.

Our $15.8ms$ (8-slice) streaming framework is faster than all prior 8-slice streaming models. After adding $6.25ms$ for a 20 Hz LiDAR to collect a \nicefrac{1}{8} slice, prior work mostly runs at 37 frames per second (FPS) \cite{polarstream}. On the other hand, our framework runs at 45 Hz. If a slice arrives every $6.25ms$, and processing by prior work takes about 20 ms \cite{polarstream}, then the current slice has to \textbf{\textit{wait}} until the previous one is processed. This is the pitfall of depending on past slices. In our case, multiple slices can be processed \textit{\textbf{simultaneously}}, which can be exploited by accelerators like FPGAs or GPUs. 

Our mAP gains are mainly from processing images. Image feature extraction is a process that is done anyway on most AV perception stacks. Image information is often indispensable, and we use a standard ResNet-50 backbone which is pre-trained on typical perception tasks. We study the effect of smaller image resolutions on the overall runtime. While images with smaller resolution lose some details, they still maintain high mAP at a much faster runtime, as shown in Table \ref{table:timing}. Fp16 seems saturated at a certain low resolution. This implies that the overhead of operating at fp16 precision overcomes its benefits at lower resolutions.

\section{Conclusion}
This study proposes a multi-modal framework for 3D detection from streaming LiDAR point cloud data. To reduce latency in detection, our approach processes LiDAR slices as they come and expands their context using camera images. Parallel encoding and independence from past slices enable further reduction in latency using hardware accelerators, especially when coupled with our proposed cropping operation. The robustness of the proposed framework to small calibration errors, missing camera images, and heavily fragmented point clouds was studied and demonstrated. Our streaming framework is shown to outperform streaming as well as representative full-scan models in terms of detection accuracy on a challenging benchmark.

\bibliographystyle{IEEEtran}
\bibliography{ref}

\end{document}